\documentclass[letterpaper, 10 pt, conference]{ieeeconf}  

\IEEEoverridecommandlockouts                              

\usepackage{amssymb}  
\usepackage{graphics} 
\usepackage{graphicx}
\usepackage[tight,footnotesize]{subfigure}
\long\def\invis#1{}
\usepackage{cite}
\usepackage{url}
\usepackage{hyphenat}
\usepackage{amsmath}
\usepackage{xcolor}
\usepackage{siunitx}

\newcommand\sect[1]{Section~\ref{#1}}
\newcommand\fig[1]{Figure~\ref{#1}}
\newcommand\tab[1]{Table~\ref{#1}}

\usepackage{algpseudocode}
\usepackage{algorithm}

\newcommand\ocomment[1]{\textcolor{blue}{JO'K:#1}}

\title{\LARGE \bf
Riverine Coverage with an Autonomous Surface Vehicle\\over Known Environments ~\vspace{-0.1in}
}

\author{Nare Karapetyan,$^\dag$ Adam Braude,$^\ddag$ Jason Moulton,$^\dag$ Joshua A. Burstein,$^{\dag\dag}$ \\
Scott White,$^{\dag\dag}$ Jason M. O'Kane,$^\dag$ and Ioannis Rekleitis$^\dag$~\vspace{-0.0in}
\thanks{$^\dag$ Computer Science \& Engineering Department, University of South Carolina,
{\tt\small [nare,moulton]@email.sc.edu,  [jokane,yiannisr]@cse.sc.edu}}%
\thanks{$^\ddag$ Computer Science Department, University of Puget Sound, {\tt\small abraude@pugetsound.edu}}%
\thanks{$^{\dag\dag}$ Earth, Ocean, and Environment Department, University of South Carolina,  {\tt\small swhite@geol.sc.edu}}%
\thanks{This work was partially supported by a SPARC Graduate Research Grant from the Office of the Vice President for Research at the University of South Carolina. The authors would also like to thank the National Science Foundation for its support (NSF 1513203, 1659514).}
}

\begin{document}

\maketitle
\thispagestyle{empty}
\pagestyle{empty}
\newcommand{\ncomment}[1]{{\leavevmode\color{blue}[#1]}}
\newcommand{\acomment}[1]{{\leavevmode\color{green}[#1]}}
\begin{abstract}
Environmental monitoring and surveying operations on rivers currently are performed primarily with manually\hyp operated boats. In this domain, autonomous coverage of areas is of vital importance, for improving both the quality and the efficiency of coverage. This paper leverages human expertise in river exploration and data collection strategies to automate and optimize these processes using autonomous surface vehicles (ASVs). In particular, three deterministic algorithms for both partial and complete coverage of a river segment are proposed, providing varying path length, coverage density, and turning patterns. These strategies resulted in increases in accuracy and efficiency compared to human performance. \invis{ We suggest a complete set of approaches for river coverage with different applications in mind. \textcolor{red}{N:[I think this is redundancy] We propose more efficient methods for surveying operations compared to current approaches in the field.}} The proposed methods were extensively tested in simulation using maps of real rivers of different shapes and sizes. In addition, to verify their performance in real world operations, the algorithms were deployed successfully on several parts of the Congaree River in South Carolina, USA, resulting in total of more than 35km of coverage trajectories in the field. \invis{\textcolor{green}{N: overall distance covered is 35.82 km in case it makes sense adding here}}
\end{abstract}
\section{INTRODUCTION}

Bathymetric surveys - surveys of the depth of a body of water, are an important tool for understanding hydro\hyp geologic processes, water resource management, and infrastructure maintenance. Since the sensor footprint of a bathymetric sensor is significantly smaller than the width of many rivers, a complete bathymetric survey of a river requires multiple boats or multiple passes. The usual method for performing coverage of a known two-dimensional area, the boustrophedon coverage~\cite{Acar2002J1,Acar2002J2}, performs poorly in tight and uneven spaces such as rivers. Fortunately, river surveyors have developed and practiced a variety of coverage techniques that are suitable for rivers. The development of these surveying/coverage strategies was guided by the property to be measured and the available resources. For example, studying sediment transfer in a fast flowing river requires sampling locally across the river.  Otherwise, by the time the surveyor returns to the same spot, the sediment will have moved significantly.\footnote{As Heraclitus said ``Everything changes and nothing remains still ~\ldots~and~\ldots~you cannot step twice into the same river.''~\cite{Heraclitus}.} On the other hand, sampling trajectories across the river results in an excessive number of turns, which is detrimental to the performance of certain sensors.  
In this paper, we address the question of how to conduct such coverage/sampling surveys using autonomous robots in order to increase the efficiency and accuracy, reduce cost, and eliminate the risks to human operators.  Three methods, \invis{spanning the complete coverage field,} each suited to a different scenario\invis{and strategy used by surveyors}, are presented: \invis{N:[the order in which the algorithms are presented is changed: first L-cover, z-cover, t-cover]}
\begin{enumerate}
    \item Complete coverage, reducing rotations. The ASV performs  longitudinal passes, traveling roughly parallel to the river shores.  This approach is particularly apt when the survey is being conducted using sensors, such as a side-scan sonar, that are sensitive to turning motions.  The method, termed L-Cover, adapts the number of passes depending on the width of the river.
	\item Limited resources surveying. This approach is appropriate when there is a limited budget of time or energy and the length of the river segment to be surveyed needs to be maximized.  In this method, termed Z-Cover, the ASV travels along the river in a zigzag pattern, turning away from the shore each time it reaches it. This allows bathymetric data reflecting the full width of the river to be sampled in a single pass.
	\item Localized complete coverage. In rivers with high flow, measurements are required across the river in a short time interval in order to monitor the bottom structure. The ASV is guided in a a lawn\hyp mowing pattern in the transverse direction, a pattern similar to one in the Boustrophedon Cellular Decomposition~\cite{Acar2002J1}. This strategy, termed T\hyp Cover, results in frequent rotations of the ASV.
	\invis{
	The third method is an alternative complete coverage approach is implemented. It will by guiding the boat across the river, performing a lawn\hyp mowing pattern in a transverse direction, in a manner quite similar to traditional boustrophedon coverage~\cite{Acar2002J1}. This strategy, which we term T\hyp Cover, may be advantageous, for example, for tracking sediment transfer.}
    \end{enumerate}

\begin{figure}[t]
  \centering
  \includegraphics[trim={1cm 1cm 2cm 1cm}, clip, width=0.9\columnwidth]{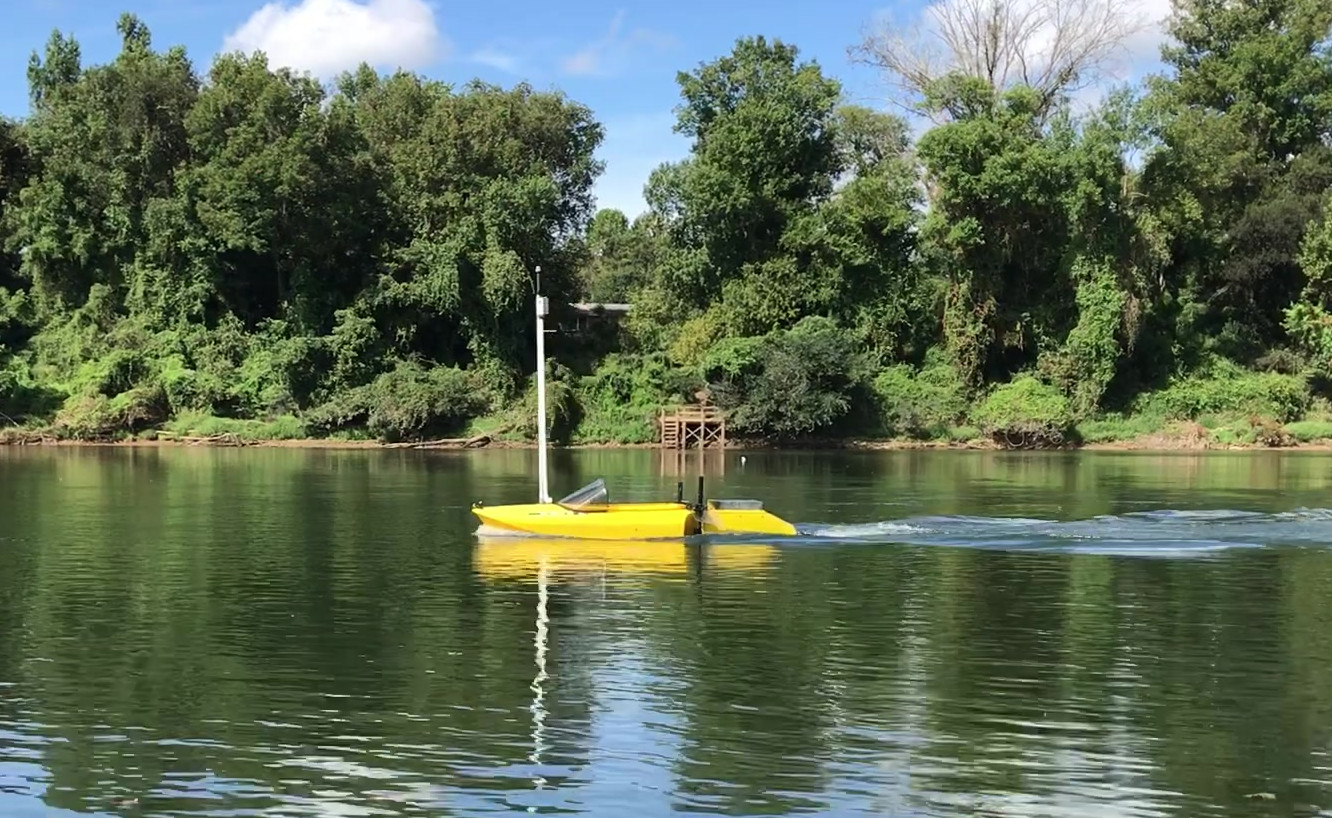}
  \vspace{-0.1in}\caption{An autonomous surface vehicle during a coverage experiment on the Congaree River near Columbia, SC, USA.\vspace{-0.3in}}
  \label{fig:beauty}
\end{figure}

The proposed complete coverage strategies (L\hyp Cover and T\hyp Cover) move along and across the river respectively, ensuring that for a fixed sensor footprint no area remains uncovered. The efficient sampling strategy ensures that the amount of information is maximized, sampling across the river multiple times, while minimizing the distance travelled. We present experiments in which all three planners were deployed on custom ASV~\cite{MoultonOceans2018} designed and fabricated in the authors' lab\footnote{The design of the ASV is publicly available on \url{https://afrl.cse.sc.edu/afrl/resources/JetyakWiki/}}.  Both quantitative assessments of the above algorithms in terms of the properties of the generated paths along with  the percentage of the region of interest covered, and qualitative results acquired from field experiments are presented. More than 35km of coverage trajectories were tested to verify the feasibility of the proposed strategies. 

In the following section we present a\invis{comprehensive} survey of related work. Next, Section~\ref{sec:methods} formally defines the problem of riverine coverage with discussions of the proposed methods.  Section~\ref{sec:results} presents both extensive simulated results and field experiments with both quantitative and qualitative analysis. Finally Section~\ref{sec:conc}, gives an overview of the proposed methods and  remarks on future work.

\invis{
\ocomment{The intro needs to be very clear about what was accomplished.  I think it's: 1. Introducing the riverine coverage problem.  2. Proposing three algorithms, called P-, T-, and Z-cover to solve it.  3. Testing those algorithms in simulation.  4. Deploying them in the field.  Is that right?  I recommend a paragraph starting with ``The contributions of this paper are...''}
}

\section{RELATED WORK}
\label{ch:litReview}

Substantial work has been done on the design and operation of autonomous surface vehicles (ASVs) in rivers. One team of researchers has shown that it is both possible and desirable to design and operate autonomous surface vehicles for the purpose of performing bathymetric surveys~\cite{ferreira2009autonomous}. Significant progress has also been made on the problem of navigating a river with an ASV \cite{snyder2004autonomous}. Additionally, another team has determined a technique for exploring and mapping a river using an unmanned aerial vehicle \cite{jain2015autonomous}. Though, to the best of our knowledge, there is no existing research on automated vehicles zig-zagging their way through rivers, the basic principle has been applied using an underwater autonomous vehicle \cite{zhang2013two}. The vehicle used repeated 135 degree turns to map an upwelling front underwater, covering 200 square kilometers over the course of five days without human intervention. Estimating the meanders of a river has also been studied by Qin and Shell~\cite{qin2017robots}, and the proposed estimator can be used for online path selection.

The problem approached in this paper is essentially a variant of the well-studied coverage problem \cite{galceran2013survey}. Of particular relevance are two works dealing with the coverage of rivers using drifters: vehicles that do not have sufficient power to travel against the current \cite{kwok2010deployment,kwok2010coverage} and another work dealing with coverage path planning for a group of energy-constrained robots \cite{sipahioglu2010energy}. \invis{\textcolor{red}{That said, this paper diverges significantly from most works relating to the coverage problem in the respect that we are actively and deliberately performing partial coverage in order to conserve resources.} Tellingly, the most recent survey of the coverage field makes no mention of cases where partial coverage is desirable \cite{galceran2013survey}.}\invis{\textcolor{green}{N: I do not think that the previous statement should be there.}} One notable work breaks from the tendency to emphasize complete coverage, instead attempting to conserve time and fuel by focusing coverage on regions of interest \cite{manjanna2016efficient}. This allowed them to create a map of a coral reef area with half the distance travelled and power used than a lawnmower-style complete coverage algorithm would have required. Another paper, in which lawnmover-style coverage is applied to a Dubins vehicle, reformulates the problem as a variant of the Traveling Salesman Problem in order to obtain an optimal solution~\cite{RekleitisIROS2017a}.

Though the above selective coverage work phrases the problem in terms of coverage, it bears kinship with the literature for \textit{informative motion planning}, that is, the problem of planning a path using limited resources in order to maximize the amount of information gained. Unfortunately informative motion planning problems are usually NP-hard optimization problems. The formulation of these problems require the definition of an information metric that can be associated with the locations or path. Since the information metric cannot be known \textit{a priori} for a real-world scenario, approximations are done using methods such as Gaussian Processes. This means that informative motion planning can be applied to practical problems, such as mapping wireless signal strength on a lake \cite{hollinger2013sampling}, understanding salinity at a river confluence, or investigating algal blooms \cite{singh2009efficient} and sampling areas with high chlorophyll~\cite{RekleitisICRA2018b}.  Despite the success of these projects, qualitative considerations involved in the formulation of our problem mean that reformulating it as an informative motion planning problem would not necessarily produce data with the desired qualities, and it would be difficult to devise an information metric that obtains the desired result.


\invis{In this work, compared to the previous works in the field, we address autonomous river coverage as a geometric problem and use the resulting patterns to justify the use of each approach in different real world scenarios. In the following section, the problem is formally defined and the different approaches are described.}

In this work, we address the autonomous coverage problem for river surveying to automate common surveying techniques used by surveyors, thus increasing the efficiency of the coverage. We formulate the riverine coverage problem as a geometric problem and use the resulting patterns to justify the use of each approach in different real world scenarios. In the following section, the problem is formally defined and the different approaches are described.
\section{PROPOSED METHODS}
\label{sec:methods}
Our objective in this paper is to automate different approaches used by river surveyors and develop more efficient planning for each of them.
We consider an ASV that was deployed with a variety of depth sensors to survey the riverbed. The ASV moves within a known environment, described by an occupancy grid map $M: \mathbb{R}^2 \to \{0, 1\}$, derived from Google satellite imagery. Values of 0 indicate the portion of the river we intend to cover, while 1 indicates locations outside that region of interest, which we treat as obstacles.
From a given starting point $v_s$, we can implicitly infer the general direction of coverage.

In this section, we describe algorithms executing three types of coverage patterns in such contexts:
\begin{itemize}
    \item Section~\ref{sec:c1} presents a pattern, termed L-cover, which moves in passes parallel to the shore.  This pattern is particularly suitable for use with a side-scanning sonar.
	\item Section~\ref{sec:c2} describes a pattern, termed Z-cover, which `bounces' between the river shores. This approach is used for performing river surveying with a single pass which is suitable for long term deployments. 
    \item Section~\ref{sec:c3} proposes a T-cover pattern, where passes made across the river, perpendicular to the shore.
\end{itemize}
These three methods differ in the length of the paths they generate, in the density of the coverage pattern, and in the number of turns needed to execute those patterns.

\invis{First we propose a solution that performs complete coverage with boustrouphedon patterns along the shore of the river. It first decomposes the river into subregions based on the width of the river. And then splits each area into passes that a robot can cover in a single pass.  Next we present an algorithm that improves on the fixed-degree zig\hyp zag approach - used mostly in manual river surveying operations. And finally, another complete coverage method uses boustrophedon patterns perpendicular to the shore of the river creating $\Gamma$-like shapes.}

\subsection{Longitudinal Coverage (L-Cover)}
\label{sec:c1}

\begin{figure}
\begin{center}
\leavevmode
\begin{tabular}{cc}
\subfigure[]{\includegraphics[width=0.2\textwidth]{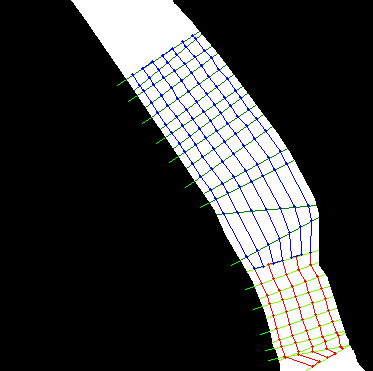}\label{fig:p1}}&
\subfigure[]{\includegraphics[width=0.2\textwidth]{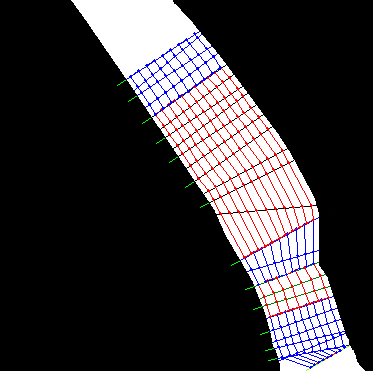}\label{fig:p2}}\\
\end{tabular}
\end{center}
\caption{An example of trajectories and clusters generated by L-Cover approach on a small section of Congaree river with different coverage density values (alternating colors mark different clusters).}
\label{fig:p_cover}
\end{figure}

Our first approach L-Cover performs coverage in a boustrophedon pattern, with passes parallel to the edges of the river. The goal of the algorithm is to split the river into subregions that can be covered with the same number of passes. The algorithm takes as input the map of the river $M$, the starting point $v_s$ and a parameter $s$ describing the desired spacing between the passes; see Algorithm \ref{alg:pcover}.  First, we identify the downriver direction (the red line in Figure~\ref{fig:explanation}) and compute an ordered list, denoted $C_{vec}$, of contour points of the shore (Line 2-3).  Next, the algorithm sequentially traverses the contour $C_{vec}$ with a step size $\Delta w$ connecting opposite edges with straight line segments, denoted $l$. $\Delta$w is the distance between each pair of segments $l_i$ and $l_{i+1}$. Then, the river is split into clusters of subregions based on the width of the river, denoted by $len()$, and the desired spacing $s$ (Lines 4-11). Any small clusters are merged with the nearest neighbor cluster that has similar width (Lines 13-15).  Finally, parallel passes are generated for each resulting cluster $Cl$ (Line 17). The resulting path $\pi$ is a list of all sequential passes from each cluster (Line 19). Examples of the results of the algorithm, with different values of $s$, are presented in Figure~\fig{fig:p_cover}.  

\begin{algorithm}
\caption{L-Cover \invis{Width Based Coverage}}
\label{alg:pcover}
\textbf{Input:} binary map of river $M$, starting point\ $v_s$\\
	\hspace*{1.20cm} spacing parameter $s$\\
	\textbf{Output:} a $\pi$ path
	\begin{algorithmic}[1]
\State $\textit{$\Delta$w} \gets \textrm{initialize}()$
\State $\textit{$C_{vec}$} \gets \textrm{getDirectionalContours}(M)$
\State $\textit{$\theta$} \gets \textrm{getDownRiverDirection}(\textit{$C_{vec}$}, \textit{$v_s$})$
\While{the end~of~the~river~is~not~reached}
\State $l \gets \textrm{getNextSegment}(\Delta w, \textit{$C_{vec}$}, l_{prev}, \theta)$
\If{$len(l) - len(l_{prev}) \le s$} 
\State insert $l$ in to $Cl_{curr}$
\Else
\State save~cluster~$Cl_{curr}$~in~$Cl_{vec}$
\State $Cl_{curr} \gets \textrm{createNewCluster}(l)$
\EndIf
\EndWhile
\For{\textbf{each} $\textit{Cl} \in \textit{$Cl_{vec}$}$}
\State Merge~$Cl$~with~closest~neighbor~within~tolerance
\EndFor
\For{\textbf{each} $Cl \in Cl_{vec}$}
\State $\textit{p} \gets \textrm{generatePasses}(\textit{Cl},\textit{s})$
\State append $p$ to $\pi$
\EndFor
\State \Return $\pi$
\end{algorithmic}
\end{algorithm}

\subsection{Zig\hyp Zag Coverage (Z-Cover)}
\label{sec:c2}

 Z-Cover partial coverage approach, is based on a zig\hyp zag pattern which aims to cover a substantial portion of the environment in a single pass along the river.
\invis{This approach is based on the fixed-angle zig\hyp zag coverage approach which aims to maximize the covered area in a single pass along the river without performing complete coverage.} The core idea of the proposed algorithm is to build a coverage path that gathers information along and across the river simultaneously. By ensuring that consecutive triangles have approximately equal areas, we ensure that the ratio of the covered areas across the river area approximately same. \invis{paths from shore to shore have approximately even lengths and the terrain is uniformly covered.}

\begin{figure}
\begin{center}
\leavevmode
\begin{tabular}{cc}
\subfigure[]{\includegraphics[height=0.17\textheight]{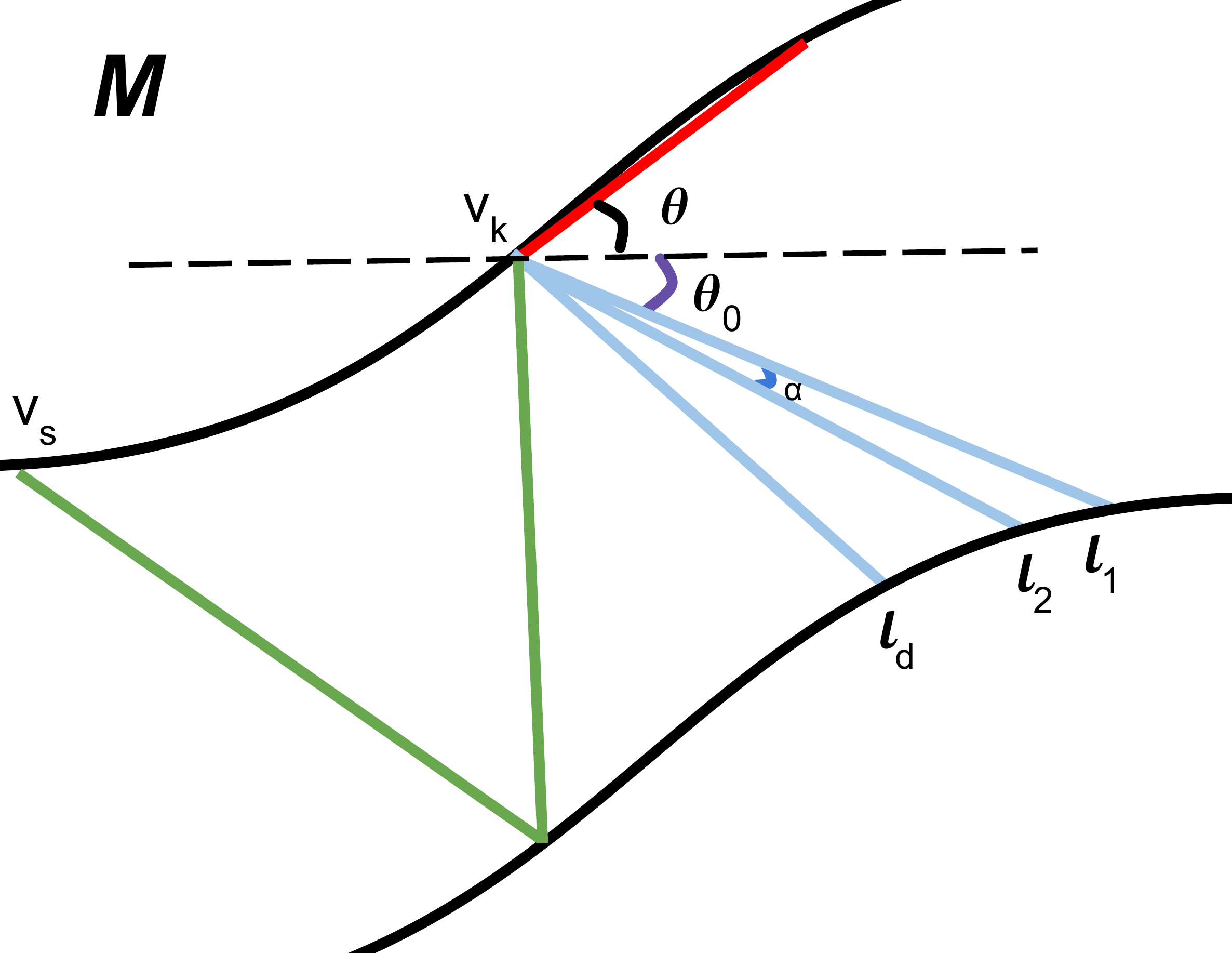}\label{fig:explanation}}&
\subfigure[]{\includegraphics[trim={0.5cm 0 0 0}, height=0.17\textheight]{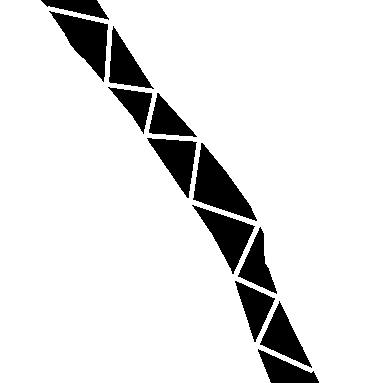}\label{fig:result}}\\
\end{tabular}
\end{center}
\vspace{-0.1in}\caption{\subref{fig:explanation} A sketch of triangle selection procedure. \subref{fig:result} A section from the result of the Equal Triangle algorithm applied on Congaree river. \vspace{-0.3in} }
				\label{fig:missioni}
				\end{figure}

Algorithm~\ref{alg:zcover} outlines the approach. It takes as input the map $M$ of the river and the starting point $v_s$. Just as in the L-Cover algorithm, the $C_{vec}$ vectors of directional contours are acquired (Line 2-3). Then, each time the algorithm searches for a next point, it does so by drawing lines from the current location towards the opposite shore. An acceptable next point is searched for among the intersections of the opposite shore with $d$ possible lines $l_1, l_2, ... l_d$ (blue lines in Figure~\ref{fig:explanation}) that form $\theta_0 + i\alpha$, $i=1,2,\ldots,d$ degree angle relative to the direction downriver. If one of these points forms a triangle with the previous two points on the path, with area within tolerance of the area of the previously selected triangle (the triangle with green edges in Figure~\ref{fig:explanation}), it will be selected. If no such point exists within $d$ intersections, the tolerance  $\Delta\epsilon$ will be increased and the algorithm will do the same search again (Lines 13-15). The tolerance $\Delta\epsilon$ is predefined and can be tuned if necessary.

\begin{algorithm}
\caption{Z-Cover\invis{Equal Triangles Path}}
\label{alg:zcover}
\textbf{Input:} binary map of river $M$, starting point $v_s$\\
	\textbf{Output:} a $\pi$ path 
	\begin{algorithmic}[1]
\State $\textit{$\theta_{0}$, d, $\alpha$, $\Delta\epsilon$} \gets \textrm{initialize}()$
\State $\textit{$C_{vec}$} \gets \textrm{getDirectionalContours}(M)$
\State $\textit{$\theta$} \gets \textrm{getDownRiverDirection}(\textit{$C_{vec}$}, \textit{$v_s$})$
\While{the end~of~the~river~is~not~reached}
\For{\textbf{each} $\textit{i} \in \textit{1,\ldots, d}$}
\State $\textit{$v_{curr}$} \gets \textrm{getIntersectionPoint}(\textit{$l_i$}, \textit{$C_{vec}$}, \textit{$\alpha$})$
\State $\textit{$p_1$, $p_2$} \gets \textrm{getPreviousTwoPoints}(\textit{$\pi$})$
\State $\textit{$S_{curr}$} \gets \textrm{computeAreaOfTriangle}(\textit{$v_{curr}$},\textit{$p_1$},\textit{$p_2$})$
\If{$|S_{curr} - S_{prev}| \le \Delta\epsilon$} 
\State append~$v_{curr}$~to~$\pi$
\EndIf
\State $\textit{break}$
\If{$i  == d$~and~$\pi$~is~empty}
\State $\textit{$\Delta\epsilon$++}$
\State $\textit{i} \gets \textit{1}$
\EndIf
\EndFor
\State $\textit{$v_{curr}$} \gets \textrm{getNextPoint}(\textit{$C_{vec}$})$ 
\EndWhile
\State \Return $\pi$
\end{algorithmic}
\end{algorithm}

\subsection{Transversal Coverage (T-Cover)}
\label{sec:c3}

Finally, we consider T-Cover, which performs a continuous lawn\hyp mower  motion pattern perpendicular to the shores of the river.  The algorithm uses the same information as L-Cover, namely the map $M$, the start location $v_s$, and the coverage spacing $s$. After acquiring directional contours, it generates passes, perpendicular to the shores, spaced by distance $s$ from each other. This is similar to covering a single cell of the Boustrophedon Cellular Decomposition, albeit the direction of the coverage varies with the river's meanders. This approach is utilized when the quantities measured change rapidly over time and the transverse profile of the river bed is required.  \invis{We include this approach to present a complete set of coverage approaches for Riverine coverage.}

\section{EXPERIMENTS}
\label{sec:results}

The performance of the proposed coverage strategies was first tested extensively on different size and shape river maps. Then, some of the generated paths were deployed both in simulation using the Stage simulator\cite{stage} and in the field to perform large scale river surveying that covered in total $35.82km$ distance.
In the latter case, the developed algorithms were deployed on the AFRL Jetyaks \cite{MoultonOceans2018}. The ASVs are equipped with a PixHawk controller that performs GPS-based waypoint navigation, a Raspberry Pi computer that runs the Robot Operating System (ROS) framework~\cite{ros} recording sensor data and GPS coordinates. In addition, different types of acoustic range finding sensors were used during deployments. \invis{In addition, two other acoustic sensors have been used to collect range data of different qualities; see \sect{sec:Sonar}. }

\invis{
\begin{figure}[ht]
\centering      
\includegraphics[width=0.8\columnwidth]{new_data/jetyak_setup.jpg}
\caption{Experimental ASV Setup.}
\label{fig:MokaiAn}
\end{figure}}

\subsection{Performance Analysis}

\begin{figure}[t]
 \centering
  \includegraphics[scale=1.8]{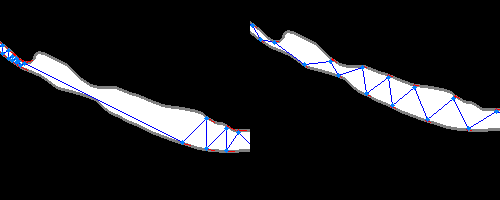}
  \caption{Contrasting two Z-Cover methods: 45 degree heuristic zig zag method (left) with the equal triangles coverage (right) described in this paper. [Note: this excerpt is 5\% of the original map]}
  \label{fig:f1}
\end{figure}

The proposed methods were tested on a set of real maps with up to $33km$ long segments of river. Because Z-cover is a partial coverage method it has been compared against a fixed-angle approach used for manual surveying operations~\cite{zhang2013two}. With the latter, boat always navigates to the opposite shore by making a fixed-angle turn relative to the near shore. The qualitative results in \fig{fig:f1} demonstrate the motivation behind the Equal Triangle Heuristic approach  for improving this operation. When automated, the fixed degree method resulted in severe overshooting and thus loss of coverage area. Meanwhile, the Equal Triangle approach ensures more even coverage. 

The primary metrics considered to evaluate performance of the coverage tasks are: 
\begin{itemize}
\item \textit{Covered Area} ($\%$), expressed as a percentage of the total area of the region of interest. For all algorithms we assume that the travel path $\pi$ has a width proportional to the spacing value $s$.\invis{, so the total area covered will be approximated by $2s$*len($\pi$).} 
\item \textit{Return Path} ($\%$), defined as the percentage of the distance traveled to return back to the starting location $v_s$ after coverage was completed over the total travel distance. This metric is especially important for large scale operations, as returning to the initial location might be time and energy consuming. 
\invis{\item \textit{Return Path Length}, defined as the distance traveled to return back to the starting location $v_s$ after coverage was completed.}
\end{itemize}

The coverage will be efficient if it will maximize covered area while minimizing the return path length. It is worth noting that in the classical \emph{coverage path planning problem,} the robot has to return to the starting position and there are areas (dead\hyp ends) where the robot enters covering and then has to traverse back resulting into double coverage. Earlier work~\cite{Xu_et_al.-2014} address this problem utilizing the Chinese Postman Problem formulation. During riverine coverage, there is only a single segment which is covered and at the end the ASV has a single return trip to the starting point, as such we do not use the total distance travelled metric as it is not informative.  

\invis{Additionally for the Z-cover we also measure the following metrics:
\begin{itemize}
\item \textit{Longest Segment (px)}: the longest distance between immediate neighbor points on the path. A high value here can be an indicator of an uneven path.
\item \textit{Segment Length Standard Deviation}: the standard deviation of the immediate neighbor points on the path. Another indicator of the path's quality.
\end{itemize}}

 \begin{table*}[ht]
  \centering
\caption{The Average results of Z-Cover, L-Cover and T-Cover approaches from simulation.\vspace{-.1in} }\label{tab:res}  
\begin{tabular}{| c | c | c | c|c |}
   \hline
   & \invis{\textit{x}} Z-Cover (Fixed-angle Heuristic) & Z-Cover (Equal Triangles) & L-Cover & T-cover \\
  \invis{ \hline
   Traveled Path Length & 7262.5 & 7594 & 21392.25 &\\}
   \hline
  \invis{ Return Path Length  & 3144.75px &	3144.25px	& 1904.02px & 3144.5px \\
    \hline}
   Return Path ($\%$) & 43.3 $\%$&	41.4$\%$ &	8.9$\%$ & 16.17$\%$ \\
  \invis{ \hline
   Longest Segment & 86.75px & 46px & NA & NA \\ 
   \hline
   Longest Segment SD \(\sigma\) & 137.5 & 50 & NA & NA \\}
   \hline
   Area Covered & 29.39\% & 31.05\% & 92.65\% & 91.42$\%$\\
   \hline
  \end{tabular}\vspace{-0.25in}
\end{table*}
In summary:
\begin{enumerate}
\item Even though T-Cover and L-Cover approaches show similar performance on completeness, when accounting for the need for a return trip, the T-Cover method is clearly outperformed by the L-Cover methods in terms of efficiency of the coverage path. 

\item The quantitative results validate the qualitative observation for differences between the Z-Cover algorithm and fixed-angle approach discussed above; see \tab{tab:res}. The Equal Triangles Method produces paths with slightly higher coverage rate.

\item T-Cover method introduces more turns in the path compared to the L-Cover. When using a side scan sonar for bathymetric mapping this can cause loss of data.
\end{enumerate}
    
\subsection{Field trials}
The main objective of the field trials is to ensure that the ASVs are able to collect adequate data when following the trajectories generated by the proposed algorithms.
We deployed an ASV to execute both the L-Cover and the Z-Cover algorithms on a 0.25$km^2$ \invis{0.4 $km^2$} area of the Congaree River, that had an average width of $91m$. For these experiments the ASV was equipped with three different Sonar sensors (see Section \ref{sec:Sonar}). Note that in this work we are assuming that the footprint of the bathymetric sensor (when side\hyp scan sonars are used) is constant and can be calculated based on the average depth of the area/river.

\begin{figure*}[ht]
\begin{center}
\leavevmode
\begin{tabular}{|c|c|c|}
\hline
\subfigure[]{\includegraphics[trim=0.5in 0.45in 0.9in 0.8in,clip,width=0.24 \textwidth]{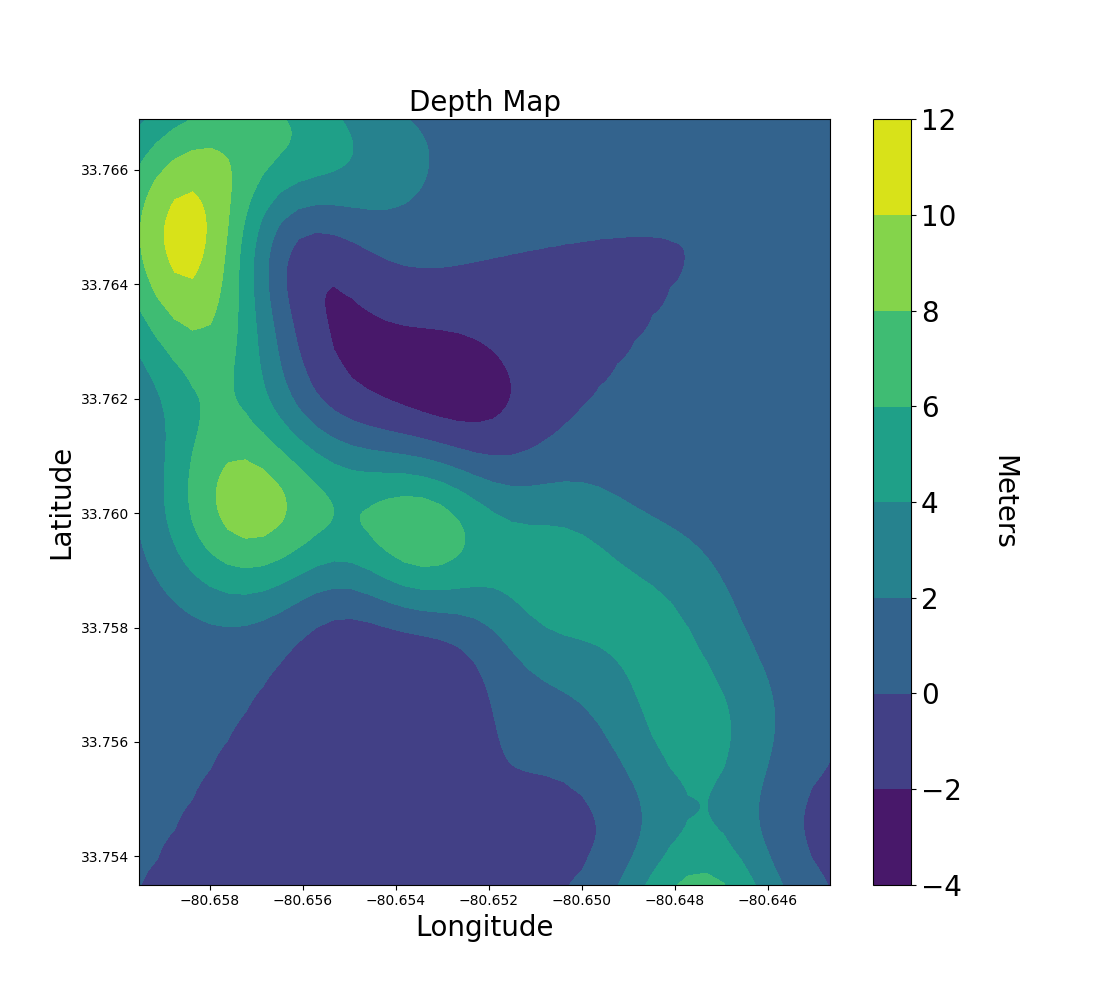}\label{fig:l1}}&
\subfigure[]{\includegraphics[trim=0.5in 0.45in 0.9in 0.8in,clip,width=0.24 \textwidth]{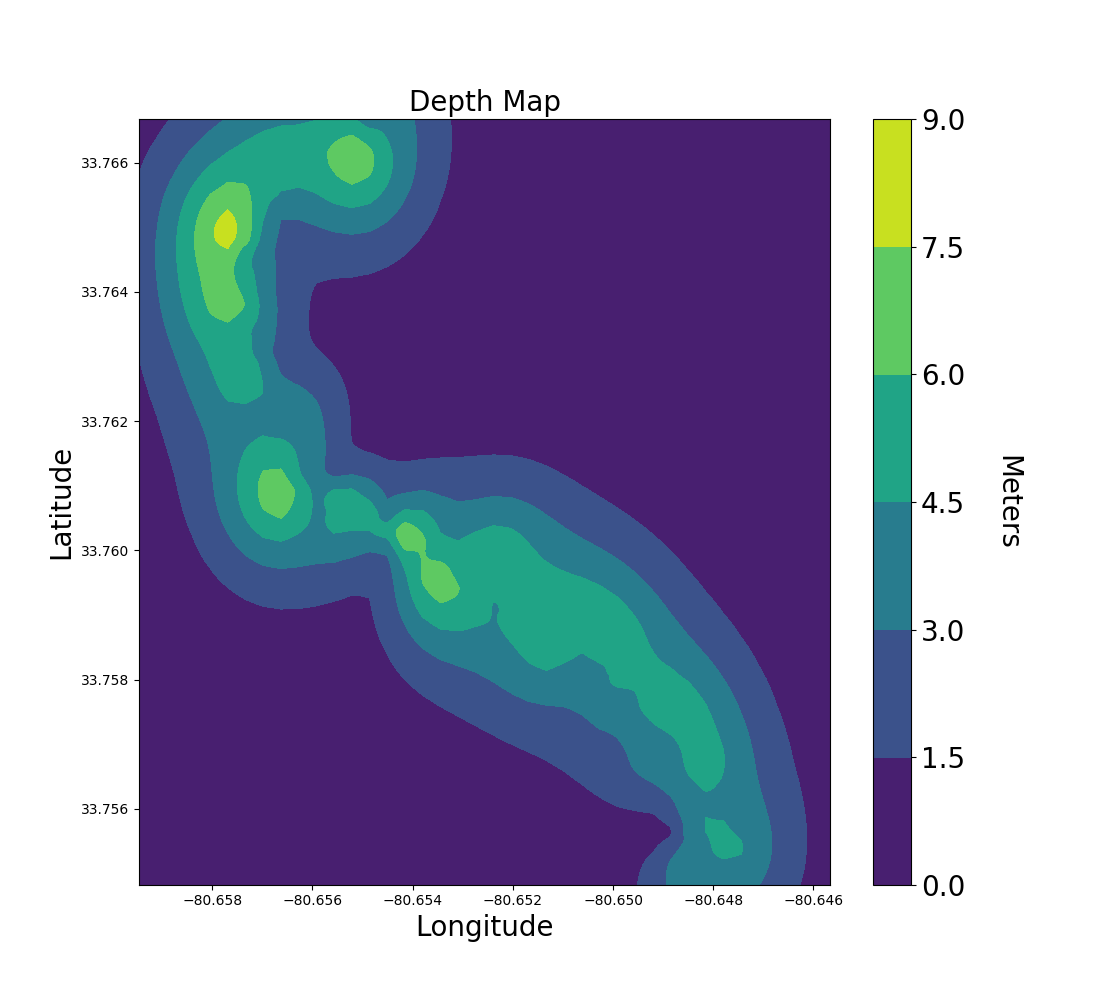}\label{fig:z1}}&
\subfigure[]{\includegraphics[trim=0.5in 0.45in 0.7in 0.8in,clip,width=0.24\textwidth]{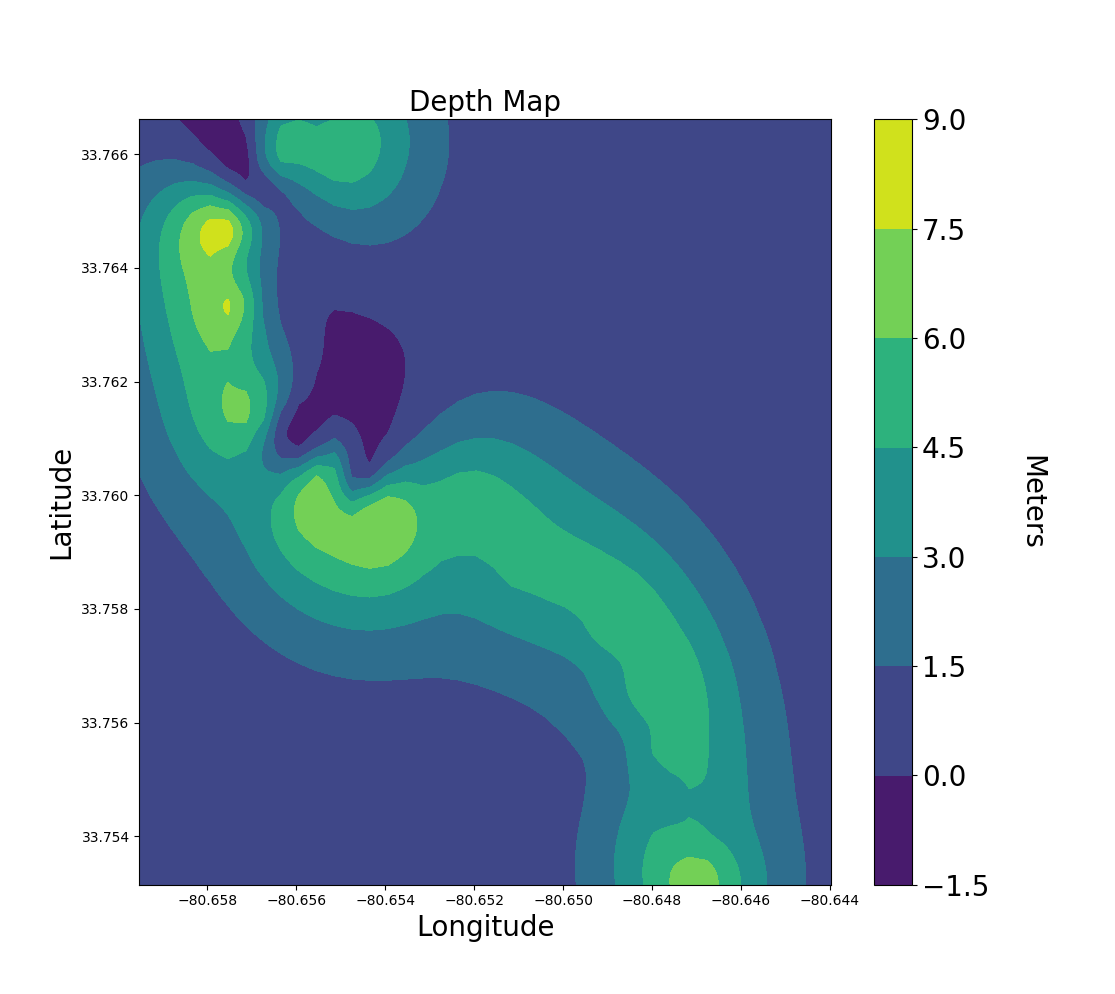}\label{fig:t1}}\\ \hline
\subfigure[]{\includegraphics[trim=0.5in 0.45in 0.9in 0.8in,clip,width=0.24 \textwidth]{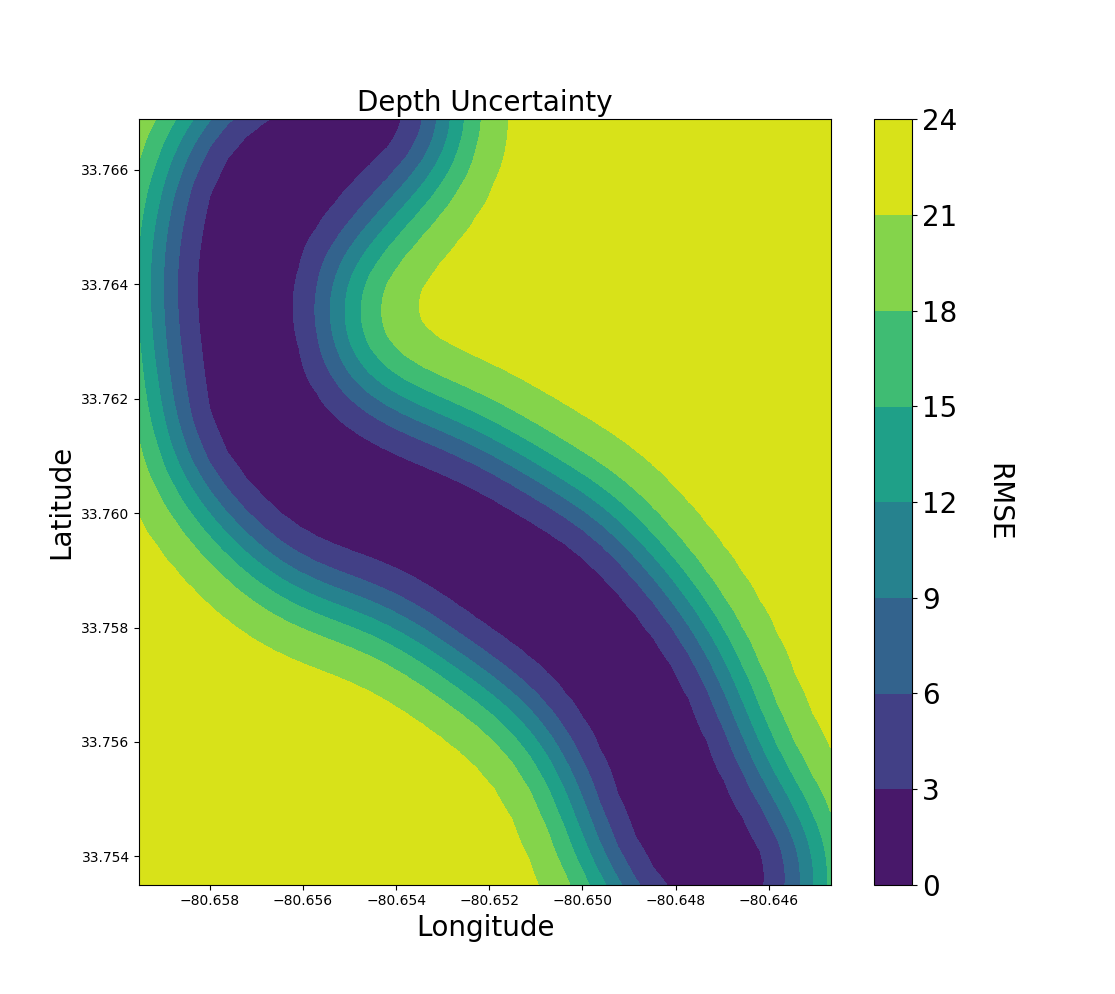}\label{fig:l2}}&
\subfigure[]{\includegraphics[trim=0.5in 0.45in 0.9in 0.8in,clip,width=0.24 \textwidth]{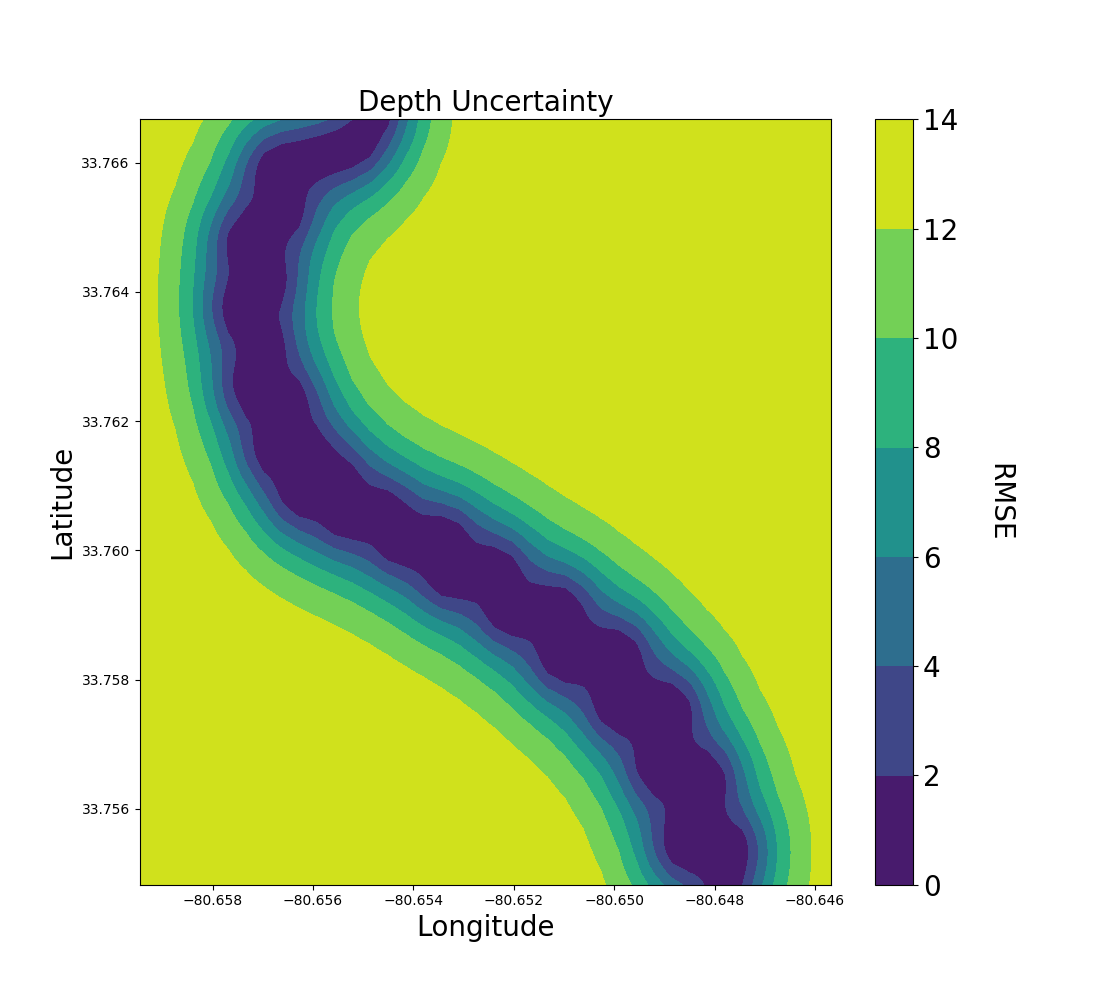}\label{fig:z2}}&
\subfigure[]{\includegraphics[trim=0.5in 0.45in 0.7in 0.8in,clip,width=0.24\textwidth]{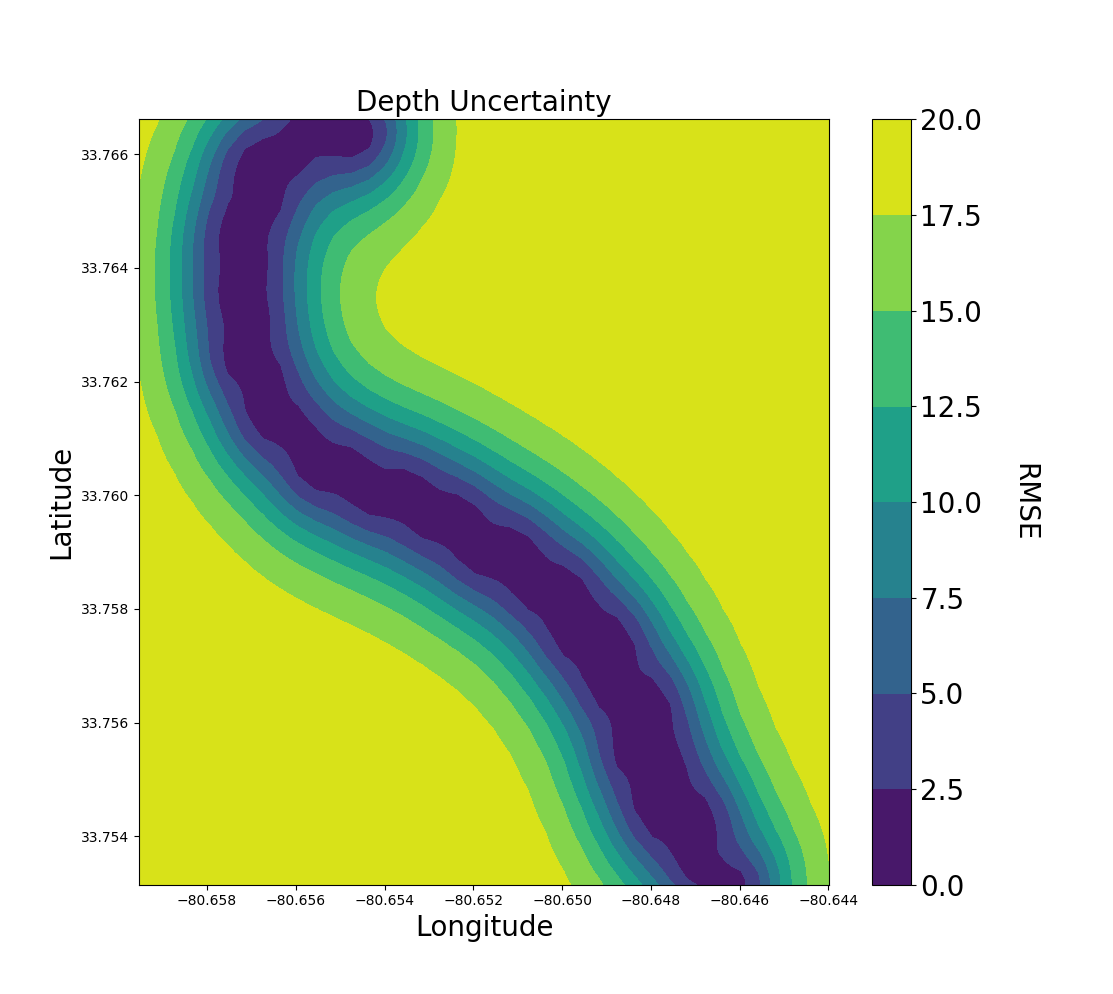}\label{fig:t2}}\\\hline
\end{tabular}
\end{center}
\caption{The depth map (in meters) of covered region and uncertainty map of selected method for that region expressed by RMSE \subref{fig:l1}, \subref{fig:l2} L-Cover, \subref{fig:z1},\subref{fig:z2} Z-Cover, \subref{fig:t1},\subref{fig:t2} T-Cover}
\label{fig:depth_rmse}
\end{figure*}

The depth measurements gathered from both experiments were used to produce a bathymetric map of the covered area utilizing a Gaussian Process (GP) mapping technique~\cite{rasmussenGP}. To evaluate the performance of both algorithms for depth map generation the uncertainty map was produced based on the root-mean-square error (RMSE). The results showed that even the operation time is longer for the L-Cover algorithm but the data collected ultrasonic range sensor is resulting in more accurate depth map. The depth map produced by data collected using L-Cover, T-Cover and Z-Cover patterns are presented in \fig{fig:depth_rmse}.

\begin{figure*}[ht]
\begin{center}
\leavevmode
\begin{tabular}{|c|c|c|c|}
\hline
\subfigure[]{\includegraphics[height=0.16\textheight]{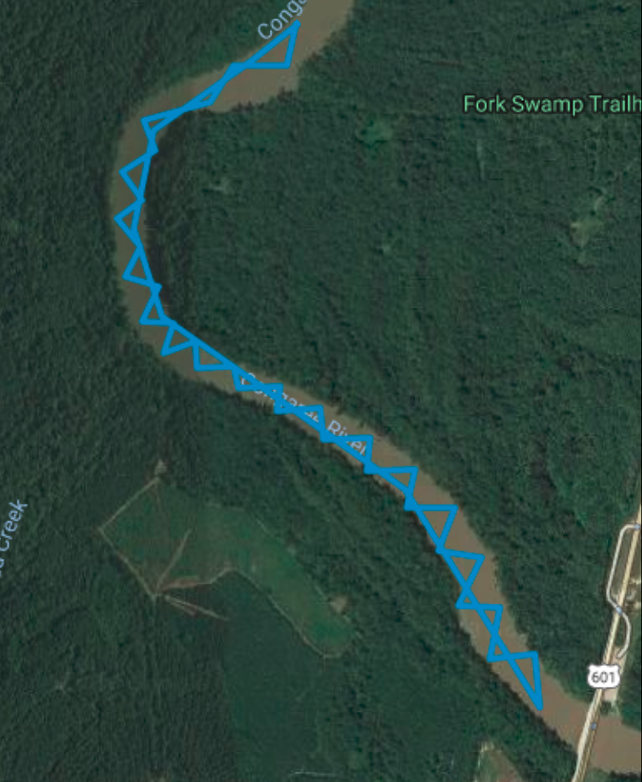}\label{fig:a1}}&
\subfigure[]{\includegraphics[height=0.16\textheight]{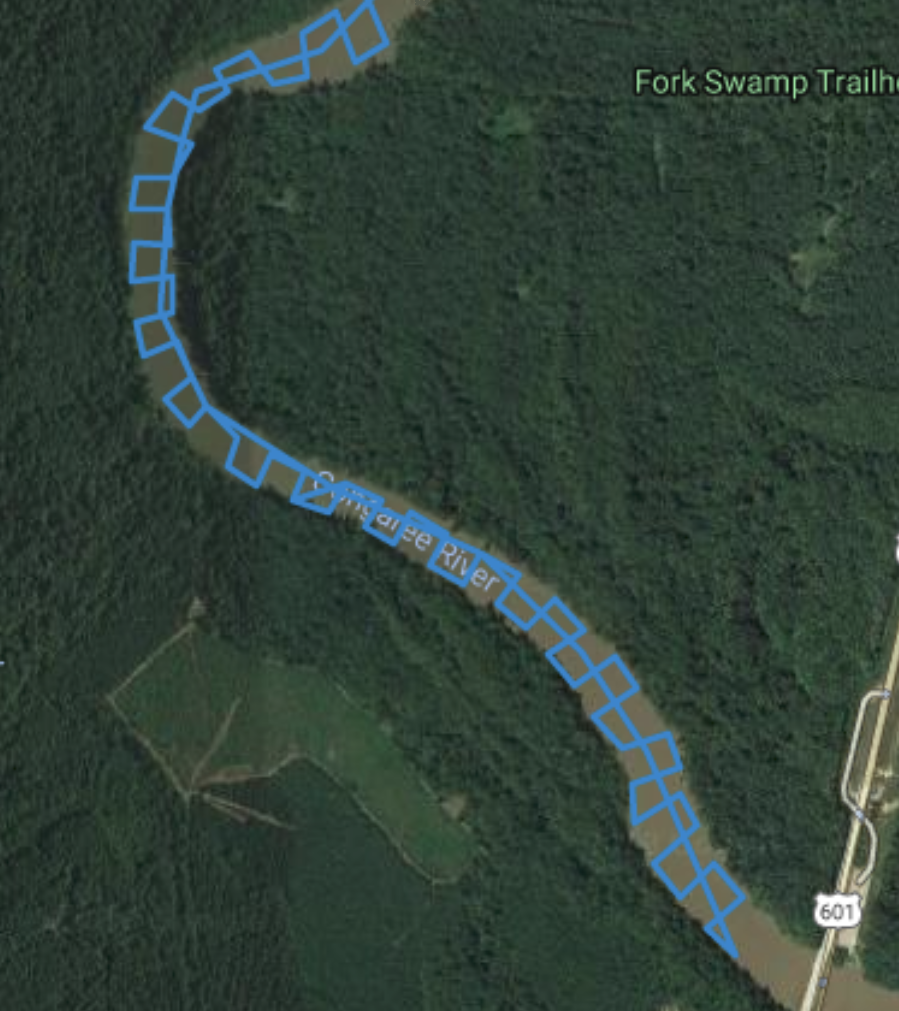}\label{fig:b1}}&
\subfigure[]{\includegraphics[height=0.16\textheight]{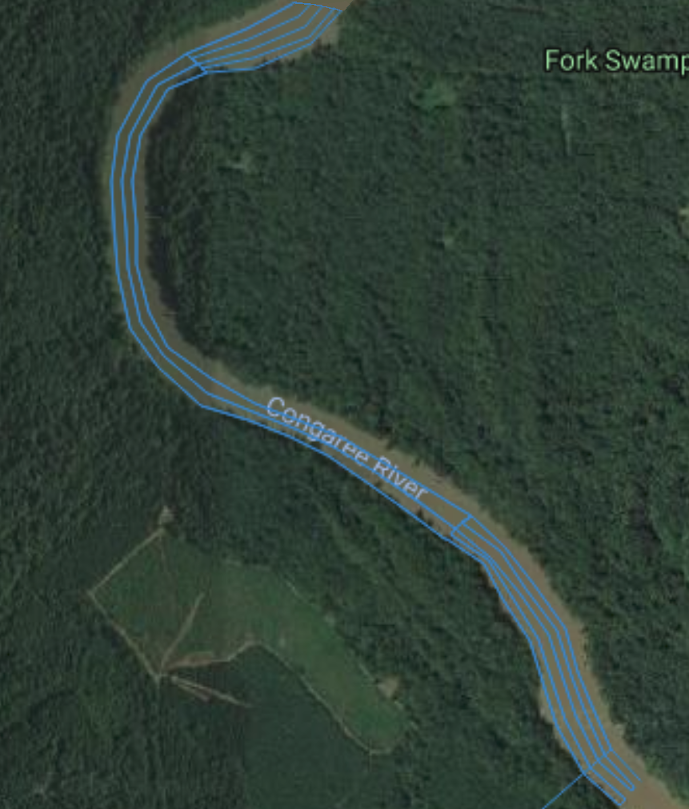}\label{fig:c1}}&
\subfigure[]{\includegraphics[height=0.16\textheight]{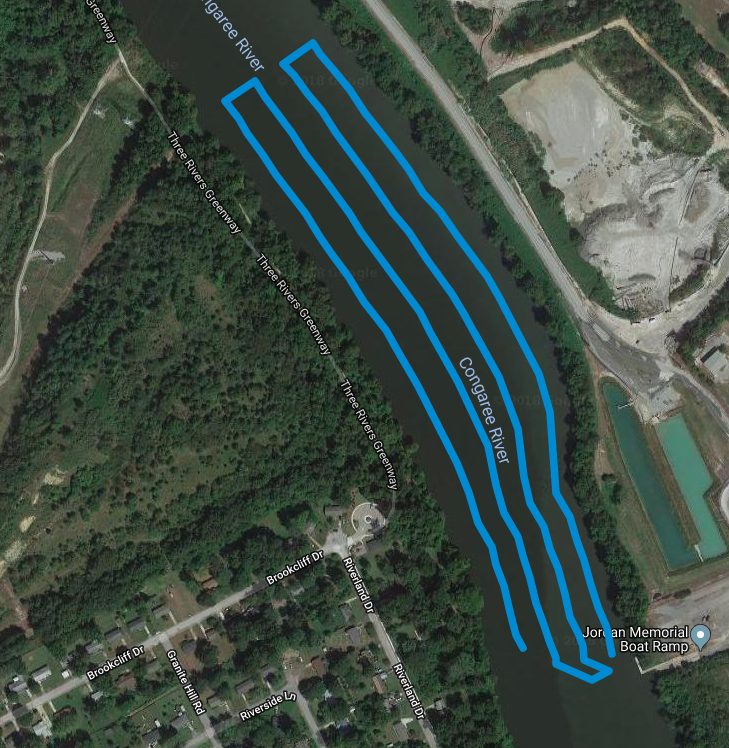}\label{fig:d1}}\\ \hline
\subfigure[]{\includegraphics[height=0.16\textheight]{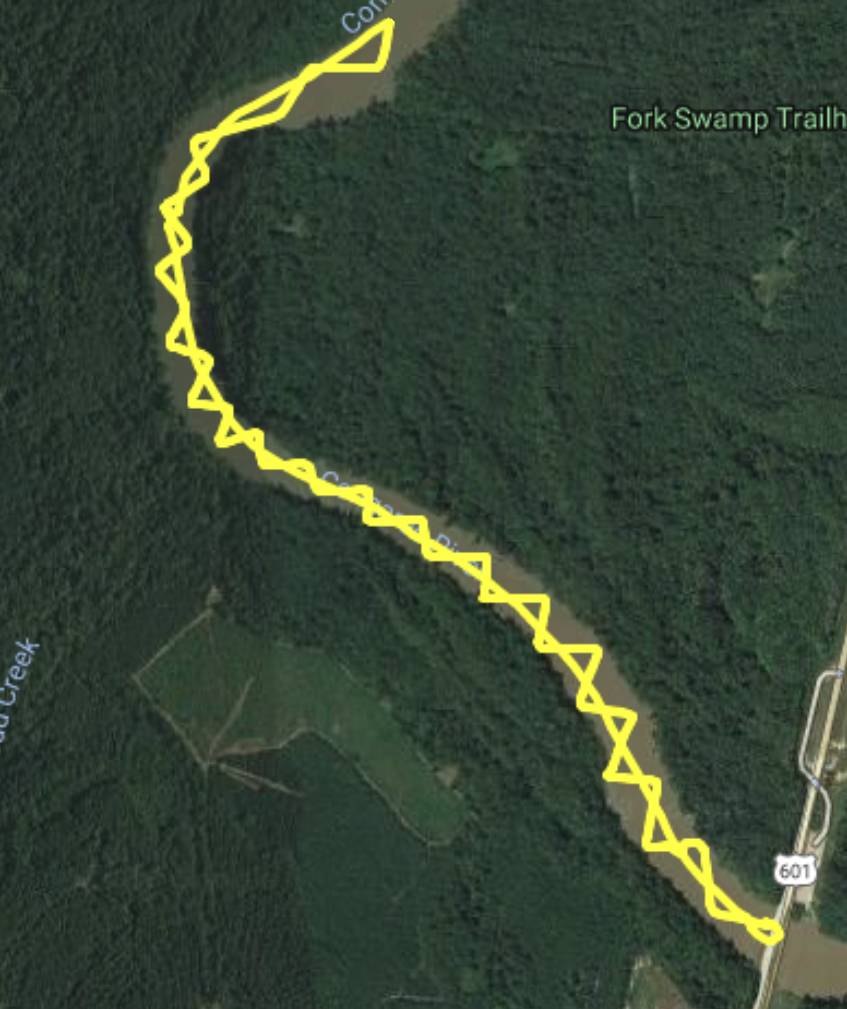}\label{fig:a2}}&
\subfigure[]{\includegraphics[height=0.16\textheight]{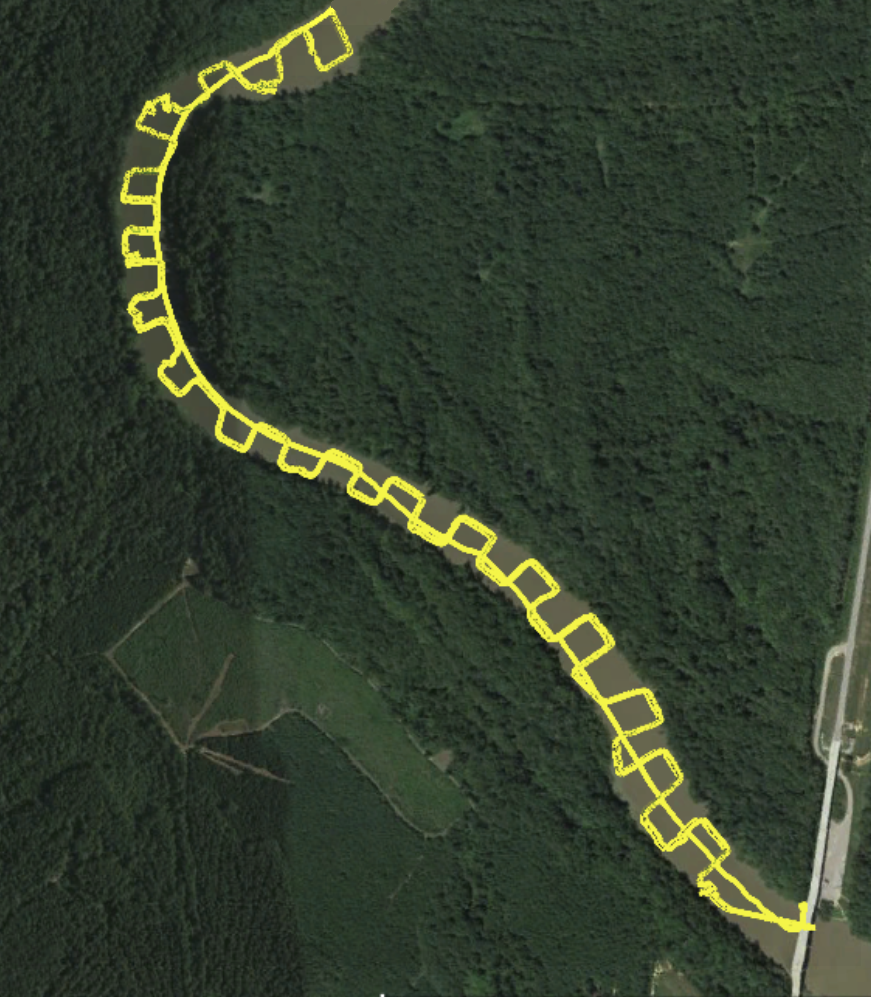}\label{fig:b2}}&
\subfigure[]{\includegraphics[height=0.16\textheight]{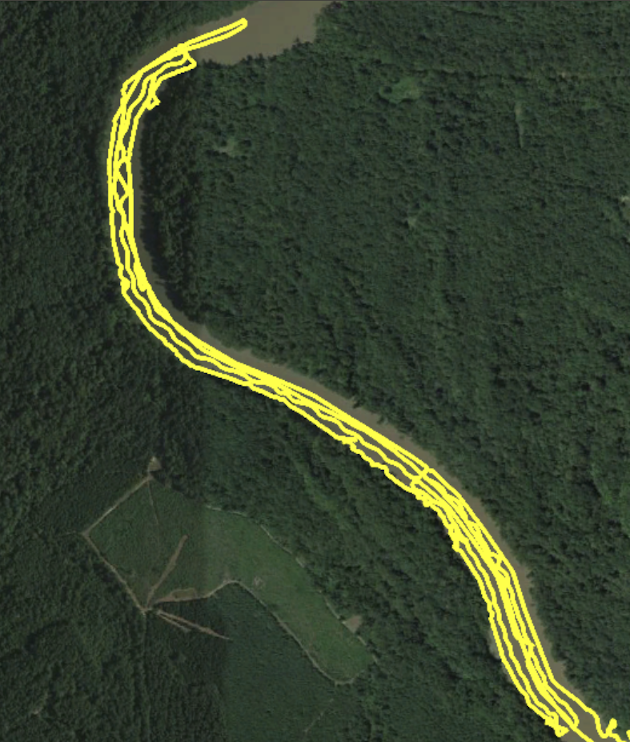}\label{fig:c2}}&
\subfigure[]{\includegraphics[height=0.16\textheight]{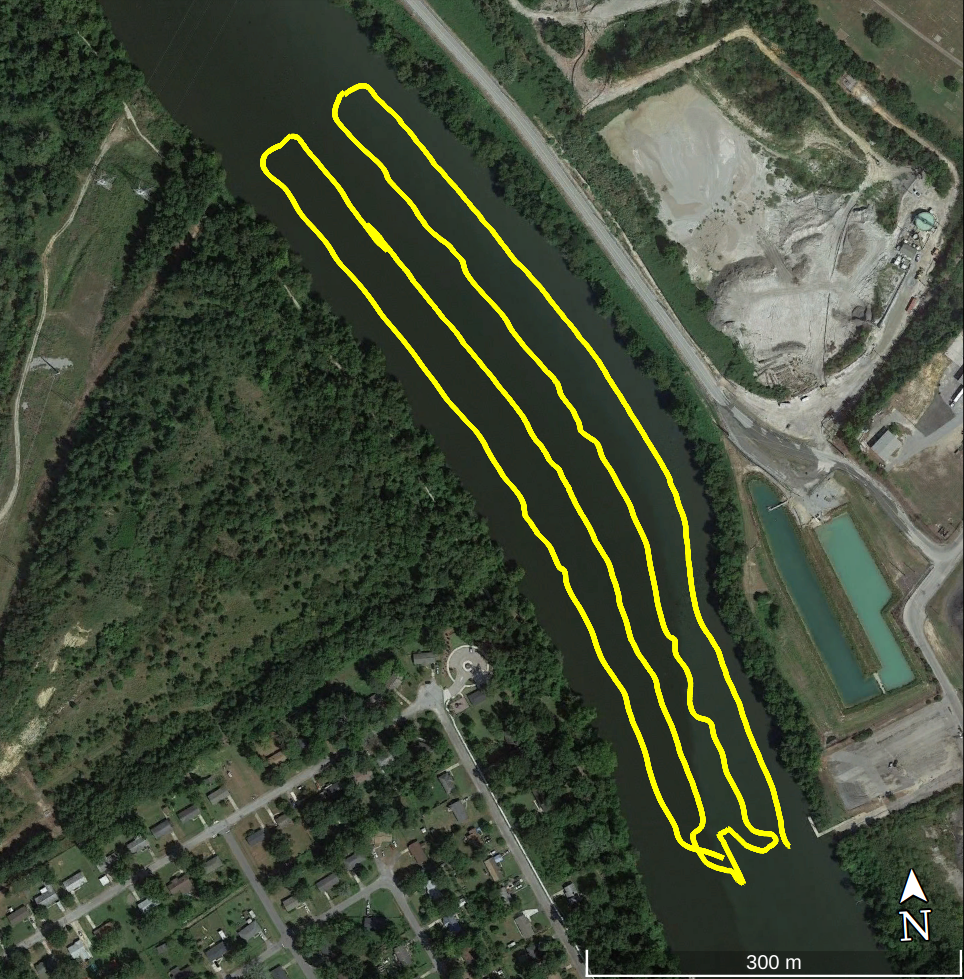}\label{fig:d2}}\\ \hline
\end{tabular}
\end{center}
\caption{Riverine Coverage on Congaree River, SC, USA. The blue paths are the ideal paths produced by the algorithms while the yellow one is boat's GPS track \subref{fig:a1},\subref{fig:a2} Z-Cover. \subref{fig:b1},\subref{fig:b2}  T-Cover. 
\subref{fig:c1}, \subref{fig:c2} L-Cover on \invis{4.3km} 2.15km long river segment. \subref{fig:d1}, \subref{fig:d2} L-Cover on 0.8 km long river segment.} 
\label{fig:mission}
\end{figure*}

\begin{figure*}[ht]
\begin{center}
\leavevmode
\begin{tabular}{cc}
\subfigure[]{\includegraphics[width=0.43\textwidth]{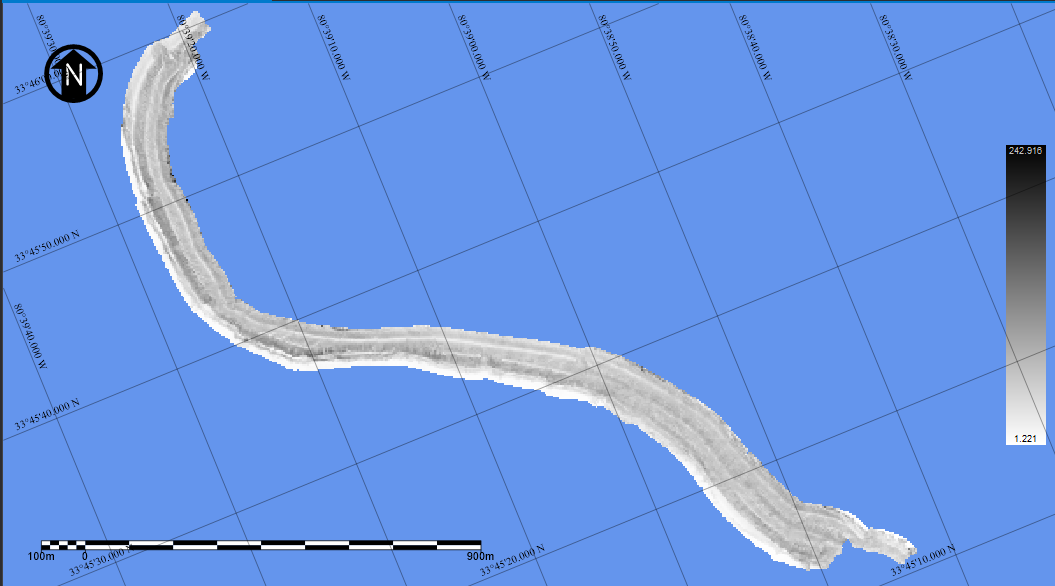}\label{fig:ab}}&
\subfigure[]{\includegraphics[width=0.43\textwidth]{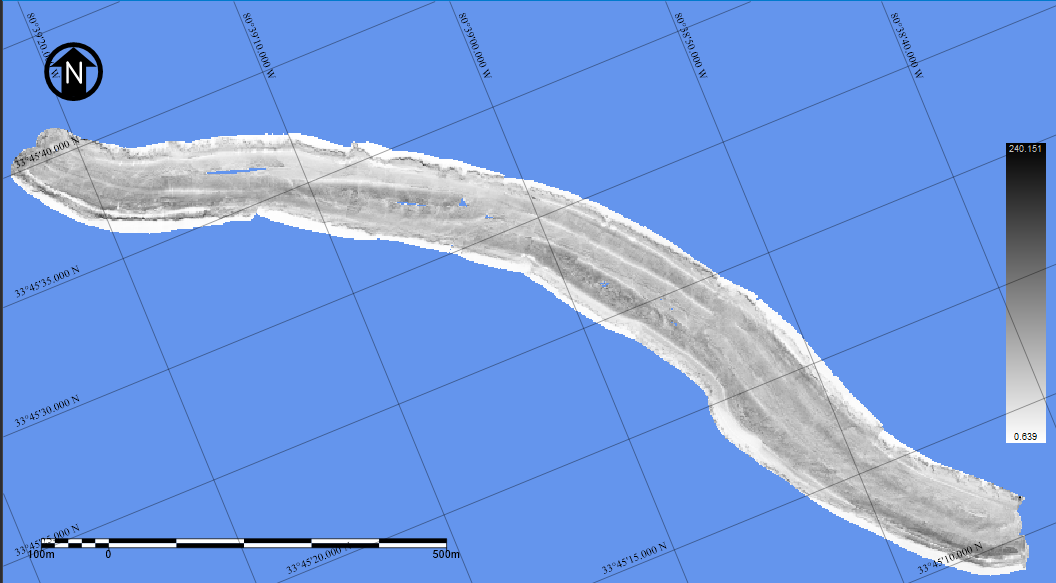}\label{fig:mb}} \\
\subfigure[]{\includegraphics[width=0.43\textwidth]{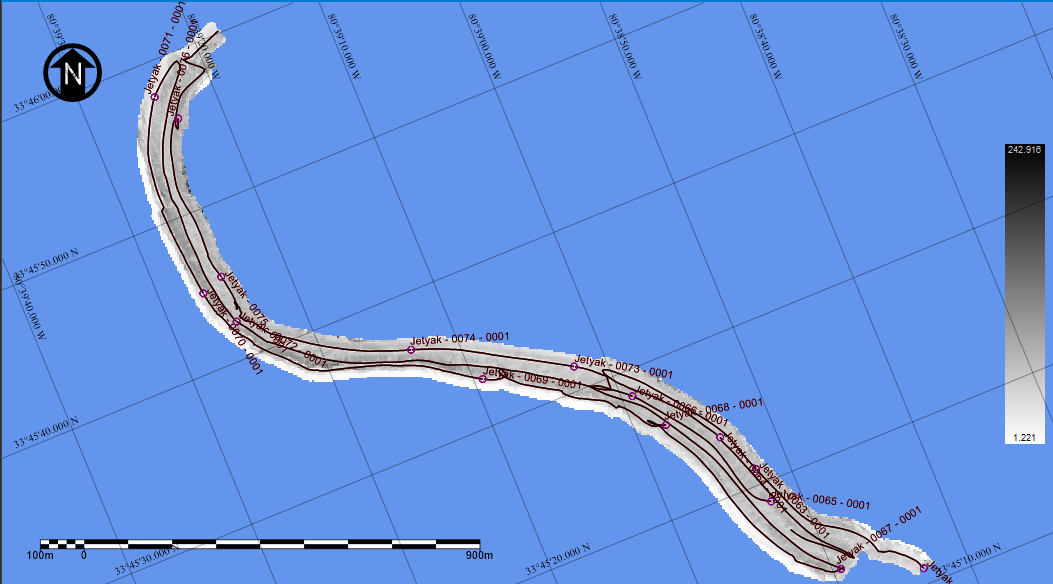}\label{fig:ab1}}&
\subfigure[]{\includegraphics[width=0.43\textwidth]{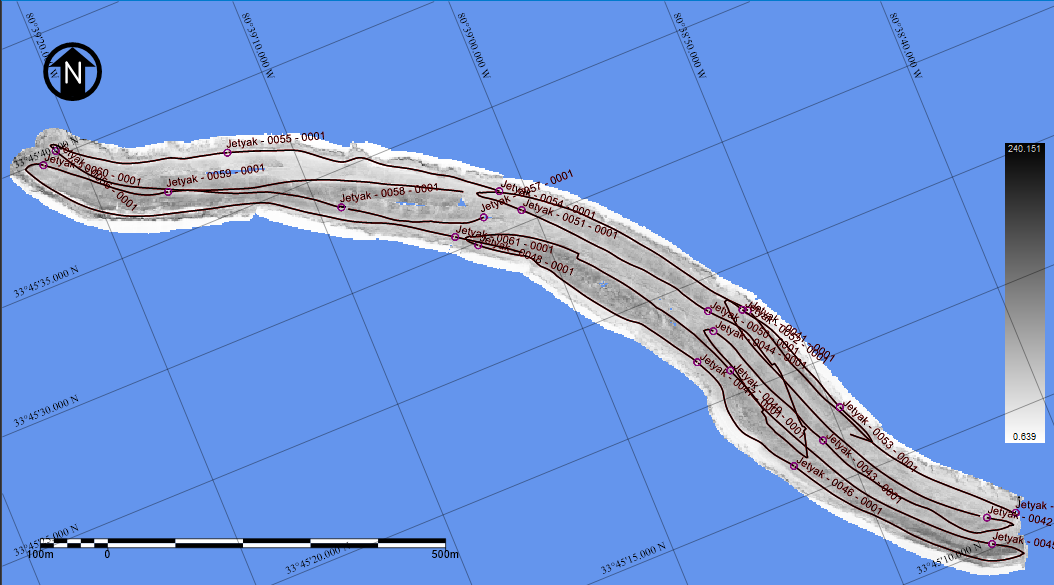}\label{fig:mb1}} 
\end{tabular}
\end{center}
\vspace{-0.2in}\caption{Backscatter image of riverbed, Congaree River. Top row shows the bathymetric map compiled from the Ping DSP data collected. Bottom row show the same bathymetric map with the ASV's path superimposed. \subref{fig:ab},\subref{fig:ab1} autonomously performing L-Cover coverage  \subref{fig:mb},\subref{fig:mb1} manually controlled.}
\label{fig:backscatter}
\end{figure*}

The boat's trajectories in Figures \ref{fig:a1}\hyp\ref{fig:d1} are closely aligned to the ideal mission plan in Figures \ref{fig:a2}\hyp\ref{fig:d2}, with small deviations caused by GPS error and environmental forces (wind, current).
The effect of those forces have been studied in our previous work~\cite{moulton2018external} and are not the subject of this work.  In addition the execution of L-Cover on the smaller region from different region of Congaree river with 0.1$km^2$ area is presented in Figures \ref{fig:d1}, \ref{fig:d2}. The resulting time and distance traveled during each experiment with actual coverage distance are presented in Table \ref{tab2}.
And finally a qualitative difference was observed when backscatter images of the riverbed were produced for both autonomous and manual coverage \fig{fig:backscatter}. It is worth noting that the manual operation trajectory is not complete compared to the path of L-Cover; see \fig{fig:backscatter}(d). The time of operation in both cases was similar (close to $2$hr) during which the autonomous operation covered a region twice the area of the manual coverage; compare \fig{fig:backscatter}(a,c) and \fig{fig:backscatter}(b,d). Moreover, the mosaicing is both complete and cleaner because of fewer overlapping tracks and odd orientations to the lines.

\begin{figure*}[ht]
\begin{center}
\leavevmode
\begin{tabular}{cc}
\subfigure[]{\includegraphics[trim=0.4in 0.45in 0.9in 0.7in,clip,height=0.2\textheight]{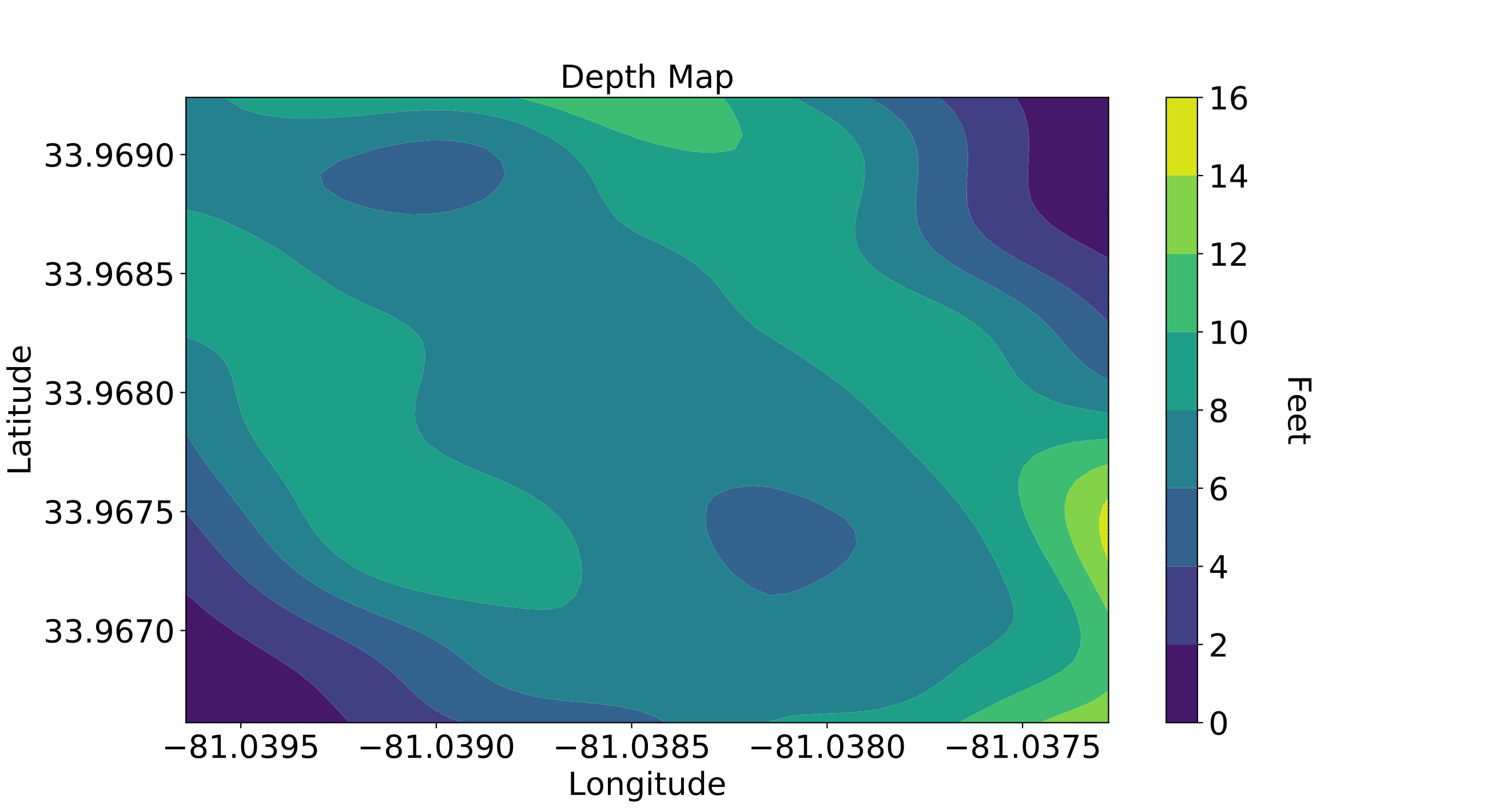}\label{fig:data1}}&
\subfigure[]{\includegraphics[height=0.2\textheight]{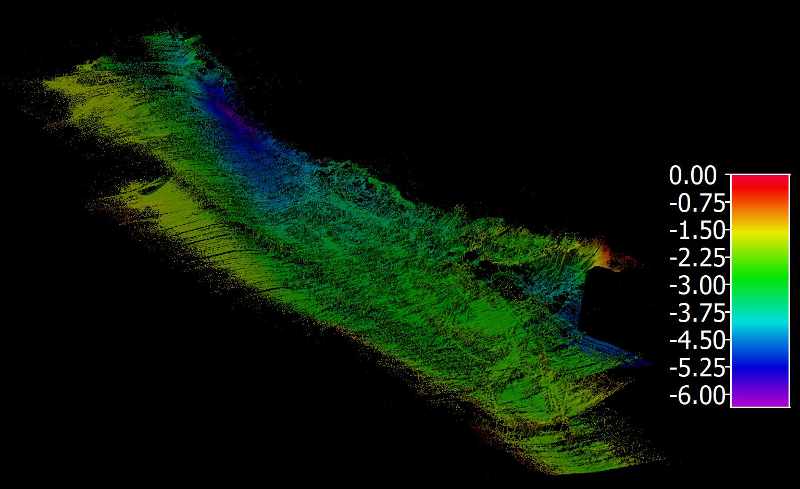}\label{fig:data3}}
\end{tabular}
\end{center}
\vspace{-0.2in}\caption{Bathymetric data collected at the Congaree River, SC, USA.  \subref{fig:data1} CruzPro depth pinger data integrated using a GP model. \invis{\subref{fig:data2} Humminbird helix 5 imaging SONAR.} \subref{fig:data3} 3DSS\hyp DX\hyp 450  side scan sonar. }
\label{fig:data}
\end{figure*}

\subsection{Riverbed Mapping}
\label{sec:Sonar}
Different acoustic sensors have been deployed over the course of the field trials in order to evaluate their performance and to consider the effect of different coverage motions to the quality of the collected data. More specifically, a CruzPro DSP Active Depth, Temperature single ping SONAR Transducer was used for the majority of the experiments; see \fig{fig:data1}. As only a single data point is collected at the time, the coverage is sparse and an integration strategy needs to be utilized as discussed above. The second sensor used was the Humminbird helix 5 chirp SI GPS G2 imaging sonar. Being a low cost, proprietary sensor, all the collected data has to be post\hyp processed.  Finally, a long range  3DSS\hyp DX\hyp 450 side scan transducer from Ping DSP\cite{ping} was deployed a limited number of times. As can be seen in \fig{fig:data3}, rotations and repeated scans do not match very well due to the sensitivity to orientation error. Acoustic data processing is beyond the scope of this paper.

\begin{table}[htbp]
\caption{Coverage time and distance results from field deployments. } 
\begin{center}
\begin{tabular}{|l|c|c|c|}
\hline
Algorithm & Z\hyp Cover  &T\hyp Cover & L\hyp Cover    \\
\hline
Time traveled & $42$m & $1$hr $09$m & $1$hr $45$m \invis{$1$hr $23$m & $2$hr $17$m & $1$hr $45$m}\\
\hline
Total Distance& 5.2km & 10km & 13.02km \invis{10.3km & 20km & 11.02km (28.6km)\footnotemark} \\ \hline
Coverage Distance & 3km & 7.3km & 11.02km \invis{6 km & 15.6km & 11.02km (23.5km)\textsuperscript{2}} \\
\hline
\end{tabular}\vspace{-0.25in}
\end{center}
\label{tab2}
\end{table}
\section{CONCLUSIONS}
\label{sec:conc}
This paper presented three strategies that perform partial and complete coverage used in different surveying scenarios. The Z-Cover algorithm shows improvements over the fixed-angle based approach used in practice by scientists for manual surveying. Both complete coverage algorithms perform boustrophedon coverage. L-Cover performs coverage parallel to the shores of the river and takes into account the width of the river for generating the passes, while T-Cover performs coverage perpendicular to the shores of the river.

The performance of the algorithms is  validated on a large number of river maps with different size and shape of contours. In addition, the algorithms were tested in simulation and in the real world. The field trials were performed on 0.25$km^2$ \invis{0.4$km^2$} and 0.1$km^2$ regions of the Congaree River. 

Taking into account the challenges encountered during field deployments, obstacle avoidance strategies must be implemented for both underwater and above water obstacles. A practical consideration is to deploy upstream, thus, in case of a failure the ASV will drift back towards the deployment site. The natural extension of this work is distributing the coverage task between multiple robots~\cite{karapetyan2017efficient, karapetyan2018multi}. Another aspect that we are interested in is planning the coverage path taking into account general knowledge about river flow based on the meanders~\cite{qin2017robots}. With this the river flow speed will be utilized to improve coverage time and energy consumption. In particular, after modeling the current in the river~\cite{moulton2018external}, path planning will associate different cost values depending on the direction of travel with respect to the direction and strength of the current.







\bibliographystyle{IEEEtran}
\bibliography{./IEEEabrv,root}


\end{document}